\def\arxivpreprint{}
\def\arxivauthors{
  Jasmine Qi, Danylo Dantsev, and Muyang Sun \\
  Indeed Inc.
}
\documentclass{article}
\ifdefined\arxivpreprint
\usepackage[preprint]{colm2026_conference}
\else
\usepackage[submission]{colm2026_conference}
\fi

\usepackage{hyperref}
\usepackage{url}
\usepackage{booktabs}
\usepackage{amsmath,amsfonts,bm}
\usepackage{graphicx}
\graphicspath{{../figures/}{figures/}}
\usepackage{multirow}
\usepackage{xcolor}
\usepackage{enumitem}
\usepackage{subcaption}
\usepackage{tikz}
\usetikzlibrary{arrows.meta,positioning}
\usepackage{lineno}
\usepackage{float}

\definecolor{darkblue}{rgb}{0, 0, 0.5}
\hypersetup{colorlinks=true, citecolor=darkblue, linkcolor=darkblue, urlcolor=darkblue}

\newcommand{\verdi}{\textsc{Verdi}}
\newcommand{\auroc}{\textsc{auroc}}

\ifdefined\arxivpreprint
\title{VERDI: Single-Call Confidence Estimation for Verification-Based LLM Judges via Decomposed Inference}
\else
\title{VERDI: Single-Call Confidence Estimation for Verification-Based LLM Judges\\via Decomposed Inference}
\fi

\providecommand{\arxivauthors}{Author names to be added before arXiv submission}
\ifdefined\arxivpreprint
\author{
  \arxivauthors
}
\else
\author{
  Anonymous Authors
}
\fi

\ifdefined\arxivpreprint
\makeatletter
\def\@maketitle{%
  \vbox{\hsize\textwidth
  \lhead{Preprint. Under review.}
  \begin{center}
    {\LARGE\bfseries \@title\par}
    \vskip 1.5em
    {\large\bfseries \@author\par}
  \end{center}
  \vskip 0.2in}}
\makeatother
\fi

\begin{document}

\ifcolmsubmission
\linenumbers
\fi

\maketitle

\begin{abstract}
LLM-as-Judge systems are widely deployed for automated evaluation, yet practitioners lack reliable methods to know \emph{when} a judge's verdict should be trusted.
Token log-probabilities, the standard post-hoc confidence signal, are unavailable for many commercial LLMs and, even when accessible, saturate above 0.999 with structured JSON output.
We introduce \verdi{} (\textbf{VER}ification-\textbf{D}ecomposed \textbf{I}nference), a method that extracts confidence from the reasoning trace a structured judge already produces, with no additional inference calls.
\verdi{} decomposes each verification-style evaluation into sub-checks and derives three structural signals: Step-Verdict Alignment, Claim-Level Margin, and Evidence Grounding Score. We combine them with Platt-scaled logistic regression.
On three public benchmarks, \verdi{} achieves \auroc{} 0.72--0.91 on GPT-4.1-mini and 0.66--0.80 on GPT-5.4-mini.
On Qwen3.5-4B/9B/27B, where answer-token logprobs are anti-calibrated (higher confidence on errors, \auroc{} 0.32--0.49), \verdi{} achieves 0.56--0.70.
We additionally validate on a production system with eight rubrics (\auroc{} 0.73--0.88 on factual rubrics), demonstrate cross-model transfer (\auroc{} 0.66--0.69), and show that a 33\,M-parameter NLI (Natural Language Inference) model provides a scalable alternative to regex extraction.
\end{abstract}

\section{Introduction}
\label{sec:intro}

LLM-as-Judge evaluation~\citep{zheng2023judging} has become the default approach for scalable text quality assessment.
Systems such as G-Eval~\citep{liu2023geval} and direct assessment prompts are used in production across content moderation, recruitment, summarization, and scientific peer review.
But a verdict alone is not enough. Practitioners need to know \emph{when to trust it}.

The standard answer relies on \textbf{token log-probabilities} (logprobs).
G-Eval~\citep{liu2023geval} weights score tokens by their probabilities; calibration methods~\citep{kadavath2022language} use logprobs to estimate prediction uncertainty.
However, logprob access is increasingly restricted~\citep{chauvin2026logprob}: as of March 2026, Anthropic Claude (all versions) provides no logprobs~\citep{anthropic2024logprobs}\footnote{Anthropic Messages API docs: \url{https://docs.anthropic.com/en/api/messages}; OpenAI Responses API docs: \url{https://platform.openai.com/docs/api-reference/responses}.}; OpenAI reasoning models hide logprobs, temperature, and top-$p$ when reasoning is enabled (GPT-5.4 exposes these controls only with reasoning effort set to \texttt{none}, which removes the reasoning traces that \verdi{} uses); and several other providers offer no documented logprob support.

Even when logprobs are available, they are often \textbf{uninformative}.
With structured JSON output (the standard for automated evaluation), we observe that 99.4--100\% of logprobs saturate to values above 0.999 across SummEval and FEVER (Table~\ref{tab:logprob_saturation}), reducing logprob-based confidence to a constant.

We propose \verdi{}, a post-hoc confidence method for structured verification judges on tasks of the form evidence + claim $\to$ analysis $\to$ verdict.
This verification pattern underlies a wide range of deployed LLM-as-Judge applications, including hallucination detection~\citep{min2023factscore}, fact-checking~\citep{thorne2018fever}, summarization consistency~\citep{fabbri2021summeval}, and attribution verification.
\verdi{} decomposes each evaluation criterion into explicit \textbf{verification sub-checks} within a single inference call: extract individual claims, verify each against a source document, and aggregate the outcomes into a verdict.
The verification-style prompt defines the judge trace; \verdi{} then scores that trace post-hoc, with no additional inference calls.
Unlike multi-call ensembles~\citep{wang2023selfconsistency} that multiply cost by 3--5$\times$, \verdi{} uses only the single original judge call.

Our contributions: (1)~Three structural signals (SVA, CLM, EGS) that extract confidence from reasoning traces with no additional inference calls.
(2)~Verification sub-checks that improve accuracy by +0.2 to +10.0pp and detect annotation staleness.
(3)~Validation on three public benchmarks (SummEval, FEVER, SciFact) across five models from two families (GPT-4.1-mini, GPT-5.4-mini, Qwen3.5-4B/9B/27B), providing confidence where logprob baselines are saturated, anti-calibrated, or unavailable, plus production validation (\auroc{} 0.73--0.88 on GPT-5.4-mini factual rubrics).
(4)~Cross-domain transferability and qualitative error analysis.
(5)~A 33M-parameter NLI model as a scalable, rubric-agnostic alternative to regex extraction.

\section{Related Work}
\label{sec:related}

\paragraph{LLM-as-Judge confidence.}
G-Eval~\citep{liu2023geval} uses token probabilities to weight evaluation scores, requiring logprob access that is increasingly unavailable or uninformative with structured JSON output.
\citet{xiong2024llmuncertainty} systematically evaluate confidence elicitation strategies and find that LLMs are consistently overconfident; \citet{tian2025overconfidence} confirm this for LLM judges specifically.
CoT-UQ~\citep{zhang2025cotuq} samples multiple reasoning paths and uses agreement as confidence; on matched 500-sample subsets, \verdi{} scores 0.667 vs.\ 0.652 on SummEval and 0.737 vs.\ 0.582 on FEVER at 5$\times$ lower cost (Section~\ref{sec:external}).

\paragraph{Trace-based uncertainty.}
\citet{kuhn2023semantic} propose semantic entropy, clustering sampled outputs by meaning to estimate uncertainty; this requires multiple samples, whereas \verdi{} uses a single trace.
\citet{devic2025tracelength} observe that trace length correlates with uncertainty; we find trace length alone achieves moderate discrimination (\auroc{} 0.56--0.57) but is improved by structural analysis (+0.10--0.16 \auroc{}).
\citet{meta2025linearprobes} use linear probes on hidden states, requiring white-box access unavailable through commercial APIs.

\paragraph{Structured evaluation.}
Chain-of-Verification~\citep{dhuliawala2024chainverification} and FActScore~\citep{min2023factscore} decompose evaluation into verifiable sub-steps to improve \emph{accuracy}.
\verdi{} shows that decomposition also produces rich \emph{confidence} signals from the existing trace.

\paragraph{Calibration and small judges.}
\citet{guo2017calibration} show that modern neural networks are poorly calibrated and propose temperature scaling; \citet{kadavath2022language} extend calibration analysis to LLMs; \citet{lin2022teaching} propose verbalized confidence; BaseCal~\citep{tan2026basecal} improves calibration across model families.
Smaller judge models~\citep{atla2025selene,bi2025judgeboard} are typically less well-calibrated, making confidence estimation even more important.

\section{Method: VERDI}
\label{sec:method}

\begin{figure}[t]
\centering
\begin{tikzpicture}[
    box/.style={draw, rounded corners, minimum height=0.8cm, align=center, font=\footnotesize},
    sig/.style={draw, rounded corners, fill=blue!8, minimum height=0.6cm, align=center, font=\footnotesize},
    arr/.style={-{Stealth[length=2mm]}, thick},
  ]
  \node[box, fill=orange!15, minimum width=1.8cm] (input) at (0,0) {Evidence\\+ Claim};
  \node[box, fill=yellow!15, minimum width=1.8cm] (judge) at (3.0,0) {Structured\\LLM Judge};
  \node[box, fill=green!10, minimum width=2.2cm] (trace) at (6.5,0) {Analysis Trace\\{\scriptsize (sub-check steps)}};

  \node[sig, minimum width=1.1cm] (sva)  at (3.8,-1.7) {SVA};
  \node[sig, minimum width=1.1cm] (clm)  at (5.3,-1.7) {CLM};
  \node[sig, minimum width=1.1cm] (egs)  at (6.8,-1.7) {EGS};
  \node[sig, minimum width=1.2cm] (surf) at (8.4,-1.7) {Surface};

  \node[box, fill=red!10, minimum width=2.0cm] (lr) at (6.1,-3.6) {Logistic\\Regression};
  \node[box, fill=purple!12, minimum width=2.0cm] (conf) at (9.2,-3.6) {Confidence\\Score};

  \draw[arr] (input) -- (judge);
  \draw[arr] (judge) -- node[above, font=\scriptsize]{single call} (trace);
  \draw[thick] (trace.south) -- ++(0,-0.3) coordinate (tb);
  \draw[arr] (tb) -| (sva.north);
  \draw[arr] (tb) -| (clm.north);
  \draw[arr] (tb) -| (egs.north);
  \draw[arr] (tb) -| (surf.north);
  \draw[thick] (sva.south)  -- ++(0,-0.55) coordinate (bL);
  \draw[thick] (clm.south)  -- ++(0,-0.55);
  \draw[thick] (egs.south)  -- ++(0,-0.55);
  \draw[thick] (surf.south) -- ++(0,-0.55) coordinate (bR);
  \draw[thick] (bL) -- (bR);
  \draw[arr]   (6.1,-2.55) -- (lr.north);
  \draw[arr]   (lr) -- (conf);
\end{tikzpicture}
\caption{\verdi{} pipeline. A structured LLM judge produces an analysis trace via a single inference call. \verdi{} extracts four signal groups from the trace and combines them via Platt-scaled logistic regression into a confidence score.}
\label{fig:pipeline}
\end{figure}
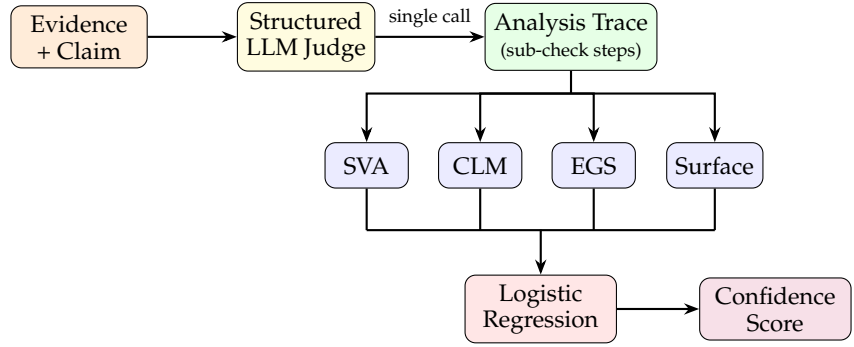

\verdi{} operates on the \textbf{analysis trace} produced by a structured LLM judge.
We assume the judge receives evidence (a document, passage, or context) and a \emph{generated} claim, that is, a factual assertion produced by an LLM rather than a statement made by a user, and then produces a step-by-step analysis culminating in a binary verdict.
This \emph{verification pattern} is common in fact-checking, consistency evaluation, and attribute verification.
Figure~\ref{fig:pipeline} illustrates the full pipeline.

\subsection{Rubric Taxonomy}
\label{sec:taxonomy}

We identify three categories of verification rubrics with distinct trace characteristics:
(1)~\textbf{Factual grounding} rubrics verify whether generated claims are supported by source documents (internal Q1--Q4; public SummEval and FEVER), producing the richest traces with per-claim evidence extraction.
(2)~\textbf{Relevance} rubrics verify whether correct content is also contextually appropriate (Q5--Q6).
(3)~\textbf{Style/integrity} rubrics check tone, grammar, bias, and overpromising (Q7--Q11), producing less structured traces.
This taxonomy predicts \verdi{} effectiveness: factual grounding rubrics produce traces that SVA, CLM, and EGS can analyze, while style rubrics yield weaker signals (Section~\ref{sec:results}).

\subsection{Verification Sub-Checks}
\label{sec:subchecks}

Rather than evaluating each rubric with a single holistic prompt (``Is the attribution accurate? Yes/No''), we decompose each into explicit verification sub-checks within a \emph{single} LLM call:
\begin{enumerate}[nosep,leftmargin=1.5em]
  \item \textbf{Claim extraction}: identify specific claims or elements to verify. If no claims are present, the sub-check terminates early with a ``nothing to verify'' verdict.
  \item \textbf{Per-claim adjudication}: verify each claim against source evidence, recording a per-claim outcome (e.g., \texttt{VERIFIED}, \texttt{FABRICATED}\footnote{\texttt{FABRICATED} denotes an LLM hallucination: the generated output contains a factual assertion not supported by the source document. It is not a judgment about the truthfulness of any user-provided content.}, \texttt{NOT\_FOUND}).
  \item \textbf{Aggregation}: combine per-claim outcomes into a final verdict with a structured analysis trace.
\end{enumerate}

\noindent
Sub-checks internalize gating logic: they handle both presence \emph{and} accuracy within a single trace, eliminating silent failures where a holistic presence-check gate incorrectly skips downstream rubrics.
A concrete comparison is in Table~\ref{tab:subcheck_example} (Appendix~\ref{app:subcheck_example}).

\subsection{Signal Definitions}
\label{sec:signals}

From the structured trace, \verdi{} extracts three core signals and a set of surface features.

\paragraph{Step-Verdict Alignment (SVA).}
For each reasoning step that contains a local conclusion (e.g., ``this claim is supported,'' ``no evidence found''), we classify the step as supporting or opposing the final verdict.
SVA is the fraction of steps aligned with the verdict:
\begin{equation}
  \text{SVA} = \frac{|\{\text{steps aligned with verdict}\}|}{|\{\text{steps with conclusions}\}|}
\end{equation}
Higher SVA indicates internal consistency; lower SVA suggests the model reached a conclusion that contradicts its own reasoning.

We implement SVA extraction in two ways:
\begin{itemize}[nosep,leftmargin=1.2em]
  \item \textbf{Regex-based}: keyword patterns match positive conclusions (``supported,'' ``confirmed,'' ``verified'') and negative conclusions (``not supported,'' ``fabricated,'' ``no evidence''). This is fast and deterministic but requires well-structured traces with consistent verdict language.
  \item \textbf{NLI-based}: a cross-encoder NLI model (\texttt{nli-MiniLM2-L6-H768}, 33M parameters) classifies whether each step entails or contradicts the verdict. This runs in under 2 seconds on CPU for a full trace and requires no rubric-specific customization.
\end{itemize}

\paragraph{Claim-Level Margin (CLM).}
CLM counts the directional balance of sub-claims:
\begin{equation}
  \text{CLM} = \frac{|\{\text{claims in majority direction}\}|}{|\{\text{total claims}\}|}
\end{equation}
A margin near 1.0 indicates unanimous support; near 0.5 indicates an even split.

\paragraph{Evidence Grounding Score (EGS).}
EGS measures how much of the analysis is grounded in quoted evidence from the source document.
We extract quoted text spans (delimited by quotation marks in the analysis trace) and fuzzy-match each span against the source document using token-level overlap with a threshold of 80\%.
EGS is the fraction of quoted spans that can be verified in the source, weighted by span length.
Hallucinated quotes (where the model fabricates evidence to support its reasoning) produce low EGS and indicate unreliable verdicts.

\paragraph{Surface features.}
We additionally extract: trace length (word count), hedging word count (``however,'' ``partially,'' ``unclear''), negation count (``not,'' ``no,'' ``never''), and quoted span count.
All signal extraction is \textbf{post-hoc}: it operates on the completed trace using deterministic parsing with no additional LLM calls.

\subsection{Confidence Scoring}

All signals are combined via \textbf{Platt-scaled logistic regression} with L2 regularization ($\lambda = 0.1$), standardized features, and 5-fold cross-validated \auroc{} with bootstrap 95\% CIs (2000 resamples).
The 7-feature model is intentionally low-capacity to prevent overfitting on datasets as small as $N=500$.
The logistic regression is fit from scratch (no external ML libraries), enabling deployment in constrained environments.

\paragraph{Intuition.}
\verdi{} rests on a simple observation: when a judge makes an error, its reasoning trace often \emph{internally contradicts} its conclusion, even when the model's token probability remains saturated near 1.0.
Token-level confidence reflects generation fluency; reasoning-level consistency reflects evidential support.  These two quantities can diverge.
SVA, CLM, and EGS each capture a different facet of this gap: SVA measures whether intermediate conclusions agree with the final verdict, CLM measures the balance of evidence, and EGS measures whether quoted evidence is real.
Their combination provides a post-hoc consistency check that is complementary to, rather than redundant with, logprob-based calibration.

\section{Experimental Setup}
\label{sec:experiments}

\subsection{Datasets}

\paragraph{Internal verification suite.}
A proprietary evaluation system with 8 binary rubrics covering attribution verification (Q1, Q2), factual accuracy (Q4), skill relevancy assessment (Q5), logical plausibility (Q7), grammar (Q8), and integrity checks (Q10, Q11).
Each rubric checks whether a factual assertion in the \emph{LLM-generated output} (e.g., ``the candidate has 5 years of Python experience'') is supported by the source document (job posting, resume, or other input context); the system evaluates GenAI output quality, not user-provided content.
$\sim$476 samples per rubric, human-labeled ground truth.

\paragraph{SummEval~\citep{fabbri2021summeval}.}
A public academic benchmark containing 1,681 source--summary pairs from CNN/DailyMail news articles across 17 summarization models.
We frame consistency evaluation as a binary verification task: ``Is this summary consistent with the source article?''
Expert consistency scores $\geq 4.3$ (out of 5) are labeled correct.
SummEval and FEVER below are included solely for reproducibility; they are not part of our internal evaluation pipeline.

\paragraph{FEVER~\citep{thorne2018fever}.}
A public fact-verification benchmark: 500 claims with gold Wikipedia evidence passages.
Binary task: does the evidence support or refute the claim?
Balanced sampling (250 SUPPORTS, 250 REFUTES).

\paragraph{SciFact~\citep{wadden2020scifact}.}
A scientific claim verification benchmark: 500 claims paired with gold evidence from biomedical abstracts.
Binary task: does the evidence support or contradict the claim?
SciFact tests domain transfer from general-knowledge (FEVER) to scientific verification.
All three public datasets are released under permissive licenses (SummEval: Apache~2.0; FEVER: CC-BY-SA~3.0; SciFact: CC-BY-NC~4.0).

\subsection{Models}

\begin{itemize}[nosep,leftmargin=1.2em]
  \item \textbf{GPT-4.1-mini}: supports logprobs, temperature control. Used for all external benchmarks and as the primary comparison model.
  \item \textbf{GPT-5.4-mini}: OpenAI reasoning model. With reasoning enabled, logprobs, temperature, and top-$p$ are unavailable; setting reasoning effort to \texttt{none} exposes these controls but removes the reasoning traces that \verdi{} scores. \verdi{} is the only confidence method among those we evaluate that requires no additional inference calls.
  \item \textbf{Qwen3.5-4B/9B/27B}: open-weight models~(Alibaba). Run locally via HuggingFace \texttt{transformers} with greedy decoding (temperature $=0$, thinking mode disabled). Answer-token logprobs extracted via a targeted forward pass over the generated sequence.
\end{itemize}

\subsection{Baselines}

\begin{itemize}[nosep,leftmargin=1.2em]
  \item \textbf{Logprob confidence}: probability of the answer token (``yes''/``no'') from the LLM's log-probabilities.
  \item \textbf{Verbalized confidence}: modified prompt that requests an explicit ``confidence: 0--100'' field in the JSON output.
  \item \textbf{Trace length}: word count of the analysis trace (prior work baseline).
\end{itemize}

\section{Results}
\label{sec:results}

We first validate \verdi{} on three public benchmarks (SummEval, FEVER, and SciFact) where all results are fully reproducible, then confirm findings on an internal verification suite covering 8 rubrics across two model families.
All AUROC values for \verdi{} are computed via 5-fold stratified cross-validation with Platt-scaled logistic regression; baseline AUROC values (logprob, verbalized, trace length) are computed directly without CV since they involve no learned parameters.
Applying Platt scaling to the logprob baseline would not recover signal: 99.4--100\% of logprob values saturate above 0.999 with structured JSON output, leaving no variance for calibration to exploit.
CoT-UQ is evaluated on 500-sample subsets due to its 5$\times$ API cost.

\subsection{Sub-Check Accuracy Gains}
\label{sec:subcheck_accuracy}

Before evaluating confidence signals, we measure the accuracy improvement from verification sub-checks over holistic prompts.
Table~\ref{tab:subcheck_acc} reports results on GPT-4.1-mini across 476 human-annotated samples.

\begin{table}[t]
\centering
\caption{Accuracy improvement from sub-check prompts vs.\ holistic prompts (GPT-4.1-mini, 476 samples per rubric). The 67 samples flagged as unreliable ground truth (GT) were identified through annotator disagreement flags in LabelStudio.}
\label{tab:subcheck_acc}
\small
\begin{tabular}{llccc}
\toprule
Rubric & Category & Holistic & Sub-Check & $\Delta$ \\
\midrule
Q1 Attrib.\ Present   & Hallucination & 96.2\% & 98.7\% & +2.5pp \\
Q2 Attrib.\ Accuracy  & Hallucination & 80.5\% & 88.1\% & +7.6pp \\
Q5 Skill Relevancy     & Relevance     & 79.2\% & 89.2\% & +10.0pp \\
Q11 Overpromising      & Integrity     & 95.8\% & 96.0\% & +0.2pp \\
\bottomrule
\end{tabular}
\end{table}

Sub-checks improve accuracy on all four rubrics, with the largest gains on rubrics requiring evidence verification: Q5 gains +10.0 percentage points and Q2 gains +7.6pp.
The integrity rubric Q11 shows only a marginal gain (+0.2pp), consistent with the observation that pattern-based rubrics produce less structured traces.

\paragraph{Stricter sub-checks reveal annotation staleness.}
When sub-checks enforce stricter criteria than the original ground truth, raw accuracy \emph{drops} but confidence calibration \emph{improves}: Q4 drops from 95.6\% to 82.1\% but gains +0.07 \auroc{}; Q7 drops from 100\% to 91.4\% but gains +0.16 \auroc{}.
This diagnostic pattern indicates \textbf{annotation staleness}: \verdi{} confidence is low on exactly these newly-flagged cases, enabling a workflow where divergences between \verdi{} scores and existing labels prioritize annotation guideline updates.
This approach flagged 67 samples where subsequent review confirmed outdated labels.

\subsection{External Benchmarks: SummEval, FEVER, and SciFact}
\label{sec:external}

We validate \verdi{} on three public benchmarks where results are fully reproducible.

\begin{table}[t]
\centering
\caption{External benchmark results (GPT-4.1-mini). \verdi{} uses no extra inference calls; CoT-UQ uses 5$\times$. Logprobs saturate above 0.999 on 99--100\% of samples with structured JSON output.}
\label{tab:external}\label{tab:logprob_saturation}
\small
\begin{tabular}{lcccc}
\toprule
Method & Calls & SummEval & FEVER & SciFact \\
\midrule
Logprob confidence      & 1$\times$ & 0.371\tiny{$\pm$0.03} & 0.523\tiny{$\pm$0.03} & 0.511\tiny{$\pm$0.01} \\
CoT-UQ~\citep{zhang2025cotuq}\textsuperscript{$\ddagger$} & 5$\times$ & 0.652\tiny{$\pm$0.05} & 0.582\tiny{$\pm$0.05} & --- \\
Trace length            & 1$\times$ & 0.561 & 0.574 & --- \\
Best single signal      & 1$\times$ & 0.675 (SVA) & 0.718 (neg.) & --- \\
\verdi{} LR ensemble    & \textbf{1$\times$} & \textbf{0.717}\tiny{$\pm$0.03} & \textbf{0.737}\tiny{$\pm$0.07} & \textbf{0.910}\tiny{$\pm$0.04} \\
\bottomrule
\multicolumn{5}{l}{\scriptsize \textsuperscript{$\ddagger$}CoT-UQ evaluated on 500-sample subsets due to 5$\times$ API cost; all other rows use full datasets.} \\
\multicolumn{5}{l}{\scriptsize Logprobs saturate above 0.999 on 99.4\% of SciFact samples, yielding near-random AUROC.} \\
\multicolumn{5}{l}{\scriptsize ``---'' = not reported (CoT-UQ not run on SciFact; trace length and best single signal omitted because SciFact uses a different prompt structure).}
\end{tabular}
\end{table}

\verdi{} achieves \auroc{} 0.717--0.910 across the three benchmarks while logprobs saturate above 0.999 on 99.4--100\% of samples.
SciFact yields the strongest result (0.910); biomedical abstracts provide more structured evidence than news articles.
CoT-UQ~\citep{zhang2025cotuq} samples 5 paths at 5$\times$ cost (\auroc{} 0.582--0.652); on the same 500-sample subsets, \verdi{} scores 0.667 vs.\ 0.652 on SummEval and 0.737 vs.\ 0.582 on FEVER.\footnote{Full-dataset SummEval \auroc{} is 0.717 ($N{=}1{,}681$); the matched-subset value (0.667, $N{=}500$) is lower due to higher fold variance.}

\subsection{Internal Verification Suite: Head-to-Head (GPT-4.1-mini)}
\label{sec:main_comparison}

\begin{figure}[t]
\centering
\includegraphics[width=\columnwidth]{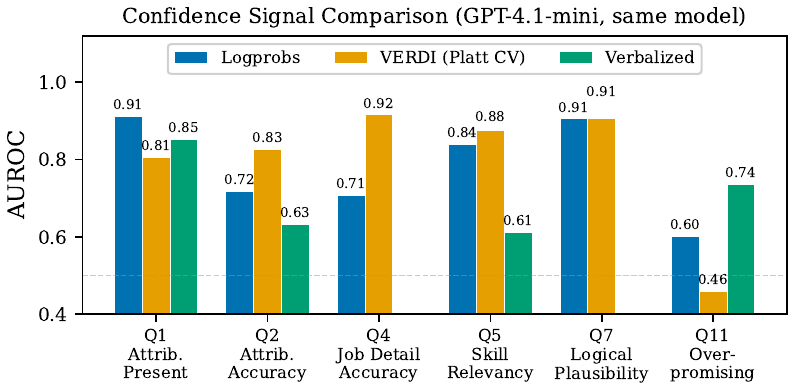}
\caption{\auroc{} comparison across methods (GPT-4.1-mini, internal data). \verdi{} wins on factual verification rubrics (Q2, Q4, Q5); logprobs win on Q1, and Q7 is a tie.}
\label{fig:auroc_comparison}
\end{figure}

Table~\ref{tab:main} and Figure~\ref{fig:auroc_comparison} compare all three confidence methods on the same model (GPT-4.1-mini).

\begin{table}[t]
\centering
\caption{Confidence signal comparison on internal data (GPT-4.1-mini). All \verdi{} numbers are 5-fold CV \auroc{}. Best per rubric in \textbf{bold}.}
\label{tab:main}
\small
\begin{tabular}{lcccl}
\toprule
Rubric & Logprob & Verbalized & \verdi{} CV & Winner \\
\midrule
Q1 Attrib.\ Present   & \textbf{0.912} & 0.851 & 0.806 & Logprob \\
Q2 Attrib.\ Accuracy  & 0.716 & 0.632 & \textbf{0.825} & \verdi{} \\
Q4 Job Detail Acc.     & 0.706 & --- & \textbf{0.915} & \verdi{} \\
Q5 Skill Relevancy     & 0.840 & 0.610 & \textbf{0.877} & \verdi{} \\
Q7 Logical Plausib.    & \textbf{0.905} & --- & 0.905 & Tie \\
Q8 Grammar             & 0.995 & --- & \textbf{1.000} & Tie \\
Q11 Overpromising      & 0.602 & \textbf{0.735} & 0.459 & Verbal. \\
\bottomrule
\end{tabular}
\end{table}

\verdi{} achieves the highest \auroc{} on Q2 (0.825), Q4 (0.915), and Q5 (0.877), the three rubrics requiring factual verification against source documents.
Logprobs win on Q1 (existence-check, 98.7\% accuracy) and tie on Q7.
Verbalized confidence wins only on Q11 (overpromising detection), a rubric where neither logprobs nor \verdi{} provide strong signals.

\paragraph{Why \verdi{} wins on factual rubrics.}
When the model makes a factual error, its intermediate reasoning often \emph{contradicts} its conclusion (low SVA), whereas logprobs remain saturated.
SVA has the largest LR coefficient on Q2 (+0.52), Q4 (+0.69), and Q5 (+0.24).

\subsection{When Logprobs Are Unavailable: GPT-5.4-mini}

GPT-5.4-mini provides no logprobs, temperature, or top-$p$. Voting and CoT-UQ baselines are impossible; verbalized confidence requires a separate prompt re-run. \verdi{} extracts confidence from the existing judge trace with no additional inference calls.

\begin{table}[t]
\centering
\caption{\verdi{} on GPT-5.4-mini (no logprobs available). 5-fold CV Platt-calibrated \auroc{}.}
\label{tab:gpt5}
\small
\begin{tabular}{lcccc}
\toprule
Rubric & N & Accuracy & \verdi{} CV & ECE \\
\midrule
Q1 Attrib.\ Present   & 471 & 99.6\% & 0.902 & 0.115 \\
Q2 Attrib.\ Accuracy  & 364 & 83.0\% & 0.726 & 0.046 \\
Q4 Job Detail Acc.     & 473 & 83.7\% & \textbf{0.879} & 0.063 \\
Q5 Skill Relevancy     & 369 & 88.3\% & 0.807 & 0.041 \\
Q7 Logical Plausib.    & 475 & 94.5\% & 0.765 & 0.032 \\
Q8 Grammar             & 475 & 99.4\% & 0.977 & 0.021 \\
\bottomrule
\end{tabular}
\end{table}

\begin{figure}[t]
\centering
\includegraphics[width=\columnwidth]{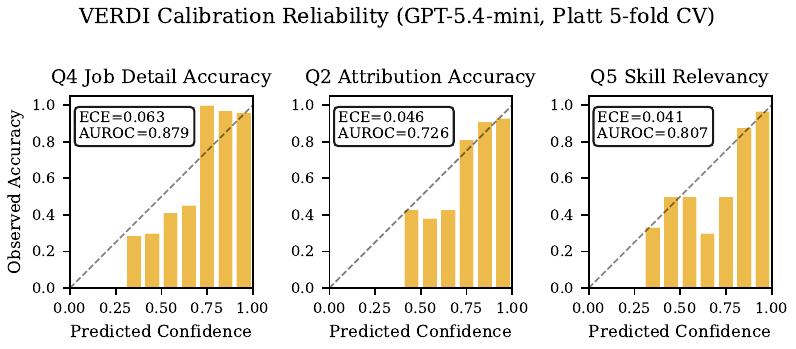}
\caption{Reliability diagram for Platt-calibrated \verdi{} on GPT-5.4-mini. Diagonal indicates perfect calibration. Q4 and Q5 show well-calibrated confidence; Q1 shows overconfidence due to the extreme class imbalance (99.6\% accuracy).}
\label{fig:calibration}
\end{figure}

\paragraph{Confidence-aware routing.}
At the optimal Youden threshold, \verdi{} flags 17--23\% of predictions for human review while catching 71--88\% of errors:
Q2 (17.6\% flagged, 71.1\% errors caught, 95.3\% retained accuracy), Q4 (22.9\%/87.7\%/97.1\%), Q5 (22.2\%/79.3\%/96.8\%).
In practice, applying a modest human review budget to the flagged subset substantially improves overall system accuracy.

\subsection{Cross-Domain Transfer}

Using a single 80/20 train/test split, FEVER-trained weights transfer best: \auroc{} 0.761 on internal ($-$0.09 vs.\ self-fit) and 0.623 on SummEval ($-$0.04).
Other source domains transfer poorly ($\leq$0.63), suggesting fact verification produces the most generalizable features (Appendix~\ref{app:ablation}).

\subsection{Signal Ablation}

The full 7-feature model provides the best or near-best performance on all datasets (Table~\ref{tab:ablation}, Appendix~\ref{app:ablation}).
Surface features alone are competitive on internal data (0.843) but do \emph{not} transfer.
SVA captures a \textbf{structural} signal (reasoning contradicts the verdict) which transfers across domains.

\subsection{NLI as Scalable Alternative}

Regex excels on well-structured traces (Q2: 0.821 \auroc{}); a 33M NLI model (MiniLM, seconds on CPU) excels on unstructured reasoning (Q5: 0.886 vs.\ 0.528 regex), providing a rubric-agnostic alternative (Table~\ref{tab:nli}, Appendix~\ref{app:nli}).

\subsection{Error Analysis}
\label{sec:errors}

\paragraph{When VERDI succeeds.}
On FEVER, 95\% of errors have saturated logprobs ($>$0.999); \verdi{} identifies 43\% through low SVA. On internal Q2, mean SVA is 0.925 for correct vs.\ 0.685 for errors (gap = 0.24), while logprob means differ by only 0.044.

\paragraph{When VERDI fails.}
On FEVER, 59\% of errors have SVA $\geq$ 0.8 (internally consistent but wrong), arising from evidence misinterpretation or threshold sensitivity on near-boundary claims.
These genuine reasoning failures are undetectable by any post-hoc signal.

\section{Discussion}
\label{sec:discussion}

\paragraph{Cost and deployment.}
\verdi{} is the only method in our comparison that provides confidence without additional calls or logprob access; G-Eval requires logprobs, verbalized confidence requires a re-run, and CoT-UQ requires 3--5$\times$ calls at temperature $>0$.
In deployment, flagging 17--23\% of predictions catches 71--88\% of errors (Section~\ref{sec:results}).

\paragraph{Beyond OpenAI models.}
Table~\ref{tab:openweight} validates \verdi{} on three open-weight Qwen~3.5 models and GPT-5.4-mini using identical benchmarks and prompts.
Qwen logprobs are anti-calibrated (\auroc{} 0.32--0.49; Appendix~\ref{app:antical}); \verdi{} achieves 0.56--0.70, a gain of +0.08 to +0.31.
GPT-5.4-mini achieves \auroc{} 0.66--0.80 with no logprob access.
Cross-model transfer (\auroc{} 0.66--0.69) supports generalization across families.

\begin{table}[t]
\centering
\caption{Multi-model validation (5-fold CV \auroc{}). Same VERDI pipeline and prompts across all models. Logprob baseline uses joint answer-token probability. Transfer trains LR on GPT-4.1-mini traces and evaluates on Qwen traces. GPT logprobs are saturated ($>$99.4\% above 0.999) or not exposed by the API.}
\label{tab:openweight}
\small
\begin{tabular}{llcccc}
\toprule
Model & Method & SummEval & FEVER & SciFact & Transfer \\
\midrule
\multirow{2}{*}{Qwen3.5-4B}
  & Logprob & 0.373 & 0.494 & --- & --- \\
  & \verdi{} LR & \textbf{0.681}\tiny{$\pm$0.07} & \textbf{0.699}\tiny{$\pm$0.13} & --- & 0.689 \\
\midrule
\multirow{2}{*}{Qwen3.5-9B}
  & Logprob & 0.417 & 0.421 & --- & --- \\
  & \verdi{} LR & 0.654\tiny{$\pm$0.08} & 0.648\tiny{$\pm$0.18} & --- & 0.693 \\
\midrule
\multirow{2}{*}{Qwen3.5-27B}
  & Logprob & 0.325 & 0.479 & --- & --- \\
  & \verdi{} LR & 0.637\tiny{$\pm$0.06} & 0.560\tiny{$\pm$0.19} & --- & 0.660 \\
\midrule
GPT-4.1-mini & \verdi{} LR & 0.717\tiny{$\pm$0.03} & 0.737\tiny{$\pm$0.07} & 0.910\tiny{$\pm$0.04} & --- \\
GPT-5.4-mini & \verdi{} LR & \textbf{0.755}\tiny{$\pm$0.02} & 0.662\tiny{$\pm$0.11} & 0.798\tiny{$\pm$0.05} & --- \\
\bottomrule
\end{tabular}
\end{table}

\paragraph{Scope and limitations.}
\verdi{} targets verification tasks (evidence + claim $\to$ verdict); it does not apply to pairwise comparison~\citep{tan2025judgebench}.
\auroc{} varies with task difficulty (0.56--0.91); when accuracy exceeds 94\%, few errors remain for any post-hoc method.
Cross-domain transfer degrades when trace conventions differ (internal$\to$SummEval: 0.52); surface features do not transfer.

\paragraph{Reproducibility.}
All public benchmark code (prompts, extraction, LR scripts, fold seeds) and the NLI model will be released upon acceptance.
Internal data cannot be shared; public benchmarks serve as the primary evaluation track.

\section{Conclusion}

We introduced \verdi{}, a method that extracts confidence from structured verification-judge traces without additional inference calls, achieving \auroc{} 0.66--0.91 on GPT models and 0.56--0.70 on Qwen on public benchmarks.
Extending trace-based confidence to open-ended generation and preference ranking remains an open problem; we release code, prompts, and all public-benchmark outputs to support reproduction.

\bibliographystyle{colm2026_conference}
\bibliography{references}

\appendix

\section{Example Sub-Check Prompt}
\label{app:prompt}

Below is an abbreviated version of the Q2 (attribution accuracy) sub-check prompt used in our experiments.
The prompt instructs the judge to extract, verify, and label each claim individually, producing the structured trace that \verdi{} analyzes.

\begin{small}
\begin{verbatim}
You are evaluating whether attributed claims
in the generated text are accurate.

Step 1 - Claim Extraction:
  Extract each specific claim attributed to
  the job seeker (e.g., "5 years of Python
  experience", "led a team of 10").
  If NO specific claims are attributed,
  answer YES (nothing to be inaccurate about).

Step 2 - Per-Claim Verification:
  For each claim, find supporting evidence in
  the resume. Label each claim:
    VERIFIED - evidence found in resume
    FABRICATED - no supporting evidence
    NOT_FOUND - claim cannot be checked

Step 3 - Aggregation:
  If ALL claims are VERIFIED -> YES
  If ANY claim is FABRICATED -> NO
  Provide final_answer and analysis as JSON.
\end{verbatim}
\end{small}

\section{Example Parsed Trace}
\label{app:trace}

The following shows how \verdi{} extracts signals from a Q2 analysis trace (abbreviated; all data below is synthetic).
\verdi{} operates entirely post-hoc on the judge's existing output trace using deterministic parsing and a local NLI model. It introduces no additional data exposure beyond what the judge already processes.
The judge evaluated three claims that the \emph{LLM-generated output} made about a candidate:

\begin{small}
\begin{verbatim}
Analysis trace:
  Claim 1: "Python developer with 5 years
   experience" -> Resume: "Python 2019-2024"
   -> VERIFIED
  Claim 2: "Led cross-functional team of 12"
   -> Resume: no mention of team leadership
   -> FABRICATED
  Claim 3: "Certified AWS Solutions Architect"
   -> Resume: "AWS SAA-C03, 2023"
   -> VERIFIED
  Final answer: NO (1 fabricated claim)

Extracted signals:
  SVA  = 2/3 = 0.67  (2 steps align with NO)
  CLM  = 2/3 = 0.67  (2 verified, 1 fabricated)
  EGS  = 2/3 = 0.67  (2 quotes verified in
                       source document)
  Hedging words: 0
  Trace length: 87 words
\end{verbatim}
\end{small}

\noindent
In this example, \texttt{FABRICATED} means the LLM hallucinated the claim ``Led cross-functional team of 12'' because the resume does not contain this information.
The low SVA (0.67) indicates that one reasoning step contradicts the final verdict: the model found supporting evidence for two claims but ruled ``NO'' based on the third.
\verdi{} assigns moderate confidence to this verdict, reflecting the mixed evidence.
A verdict with SVA~$=$~1.0 (all steps unanimous) would receive higher confidence.

\section{Rubric Descriptions}
\label{app:rubrics}

Table~\ref{tab:rubric_descriptions} provides paraphrased descriptions of all eight evaluation rubrics.
Each rubric produces a binary verdict (pass/fail) based on a structured analysis of the generated output against the source document.
Rubric IDs (Q1--Q11) follow the internal numbering; gaps reflect rubrics not included in this study.

\begin{table}[H]
\centering
\caption{Paraphrased rubric descriptions. Category reflects the taxonomy in Section~\ref{sec:taxonomy}. Trace richness indicates expected signal quality for \verdi{}.}
\label{tab:rubric_descriptions}
\small
\begin{tabular}{llp{4.8cm}l}
\toprule
ID & Category & Description (paraphrased) & Trace \\
\midrule
Q1 & Factual & Checks whether the generated output contains any attributed factual claims. Acts as a presence gate: if no claims are present, downstream accuracy rubrics are trivially satisfied. & Low \\
Q2 & Factual & Checks whether each factual claim in the generated output is supported by evidence in the source document. Per-claim verification produces a structured trace with individual outcomes. & High \\
Q4 & Factual & Checks whether specific details (e.g., dates, quantities, qualifications) in the generated output match the corresponding information in the source document. & High \\
Q5 & Relevance & Checks whether the content selected by the generated output is contextually appropriate, that is, the right information for the task, not just factually correct information. & Medium \\
Q7 & Integrity & Checks whether the generated output is logically coherent and free of internal contradictions or implausible inferences. & Low \\
Q8 & Integrity & Checks whether the generated output meets basic writing quality standards (grammar, spelling, fluency). & Low \\
Q10 & Integrity & Checks whether the generated output avoids inappropriate editorial framing, unsubstantiated characterizations, or bias in presentation. & Low \\
Q11 & Integrity & Checks whether the generated output avoids exaggeration, unsupported superlatives, or promises that go beyond what the source document supports. & Low \\
\bottomrule
\end{tabular}
\end{table}

\noindent
Trace richness reflects the depth of structured reasoning each rubric elicits.
\textbf{High}: the sub-check produces per-claim extraction, evidence matching, and individual verification outcomes; SVA, CLM, and EGS are all informative.
\textbf{Medium}: the sub-check requires contextual judgment beyond simple evidence matching, producing moderately structured traces.
\textbf{Low}: the rubric relies on pattern-based assessment without per-claim decomposition; surface features (hedging, trace length) dominate over structural signals.

\section{Holistic vs.\ Sub-Check Example}
\label{app:subcheck_example}

\begin{table}[H]
\centering
\caption{Holistic vs.\ sub-check evaluation for Q2 (attribution accuracy). The sub-check decomposes the task into per-claim verification, producing a structured trace that \verdi{} can analyze.}
\label{tab:subcheck_example}
\small
\begin{tabular}{p{0.45\columnwidth}p{0.45\columnwidth}}
\toprule
\textbf{Holistic prompt} & \textbf{Sub-check prompt} \\
\midrule
``Are the attributed claims accurate? Provide a brief analysis.'' &
``Step 1: Extract each specific claim. Step 2: For each, find evidence in the resume and mark VERIFIED or FABRICATED. Step 3: Aggregate.'' \\
\midrule
\emph{Output}: ``Yes. The claims appear consistent with the resume.'' (no per-claim detail) &
\emph{Output}: ``Claim 1: `5 years Python' -- resume says `Python 2019--2024' -- VERIFIED. Claim 2: `Led team of 10' -- no mention -- FABRICATED. Final: NO (1/2 verified).'' \\
\bottomrule
\end{tabular}
\end{table}

\section{Signal Ablation and Transfer Details}
\label{app:ablation}

\begin{table}[H]
\centering
\caption{Feature subset ablation (5-fold CV LR \auroc{}).}
\label{tab:ablation}
\small
\begin{tabular}{lcccc}
\toprule
Subset & Int.\ 4.1 & Int.\ 5.4 & SummEval & FEVER \\
\midrule
All (7 features)    & 0.832 & 0.754 & 0.642 & 0.750 \\
SVA only            & 0.698 & 0.628 & 0.562 & 0.529 \\
Surface only        & 0.843 & 0.717 & 0.629 & 0.690 \\
VERDI structural    & 0.715 & 0.689 & 0.565 & 0.704 \\
SVA + surface       & 0.834 & 0.740 & 0.645 & 0.751 \\
\bottomrule
\end{tabular}
\end{table}

\begin{table}[H]
\centering
\caption{Cross-domain LR transfer \auroc{}. Diagonal: self-fit on a single 80/20 train/test split (differs from 5-fold CV in Table~\ref{tab:external}). Off-diagonal: zero-shot transfer.}
\label{tab:transfer}
\small
\begin{tabular}{lccc}
\toprule
Train $\backslash$ Test & Internal & SummEval & FEVER \\
\midrule
Internal  & 0.855 & 0.517 & 0.631 \\
SummEval  & 0.574 & 0.662 & 0.570 \\
FEVER     & \textbf{0.761} & \textbf{0.623} & 0.796 \\
\bottomrule
\end{tabular}
\end{table}

\begin{figure}[H]
\centering
\begin{subfigure}[t]{0.48\columnwidth}
  \centering
  \includegraphics[width=\linewidth]{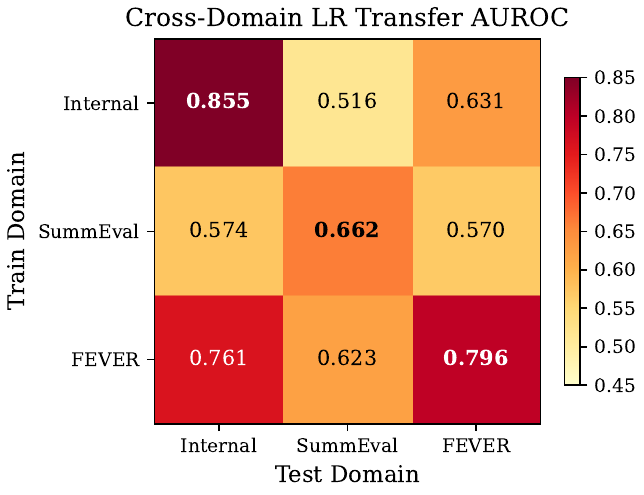}
  \caption{Cross-domain transfer.}
  \label{fig:transfer}
\end{subfigure}\hfill
\begin{subfigure}[t]{0.48\columnwidth}
  \centering
  \includegraphics[width=\linewidth]{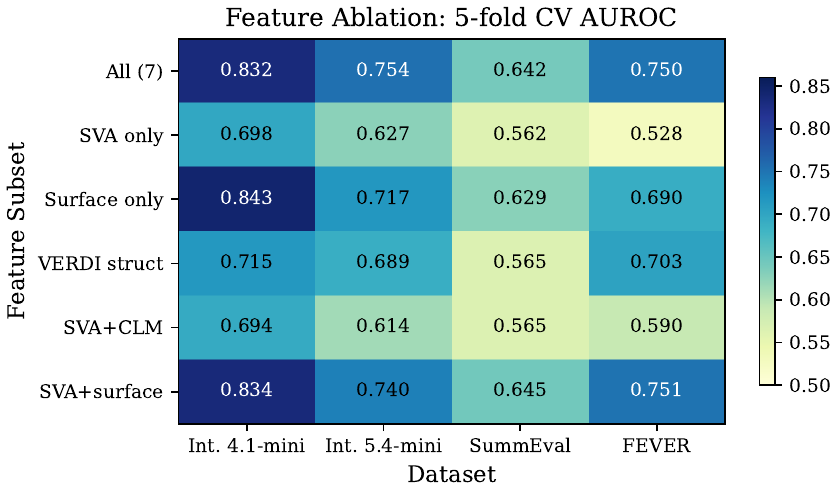}
  \caption{Feature ablation.}
  \label{fig:ablation}
\end{subfigure}
\caption{(a)~Cross-domain transfer \auroc{}. FEVER-trained weights transfer best. (b)~Feature subset ablation heatmap.}
\label{fig:heatmaps}
\end{figure}

\section{Prompt-Style Robustness: Holistic vs.\ Sub-Check}
\label{app:holistic}

Table~\ref{tab:holistic_ablation} compares \verdi{} \auroc{} when applied to traces from two prompt styles: (1)~the sub-check prompt used throughout the paper, which decomposes evaluation into numbered verification steps, and (2)~a holistic prompt that asks the model to ``analyze and explain your reasoning step by step'' without imposing structure.
\verdi{} achieves comparable AUROC under both prompt styles, demonstrating that the method does not depend on a specific prompt format.

\begin{table}[H]
\centering
\caption{Prompt-style robustness (GPT-4.1-mini, $N=500$, 5-fold CV \auroc{}). \verdi{} works on traces from both structured sub-check and unstructured holistic prompts.}
\label{tab:holistic_ablation}
\small
\begin{tabular}{lcc}
\toprule
Prompt Style & SummEval & FEVER \\
\midrule
Sub-check (structured)  & 0.709\tiny{$\pm$0.02} & 0.722\tiny{$\pm$0.07} \\
Holistic (unstructured) & 0.754\tiny{$\pm$0.08} & 0.658\tiny{$\pm$0.10} \\
\bottomrule
\end{tabular}
\end{table}

\noindent
The sub-check prompt is slightly better on FEVER ($+0.064$) and the holistic prompt slightly better on SummEval ($+0.045$); both differences fall within one standard deviation, indicating that performance is robust to prompt structure.

\section{Logprob Anti-Calibration on Qwen Models}
\label{app:antical}

\begin{figure}[H]
\centering
\includegraphics[width=\columnwidth]{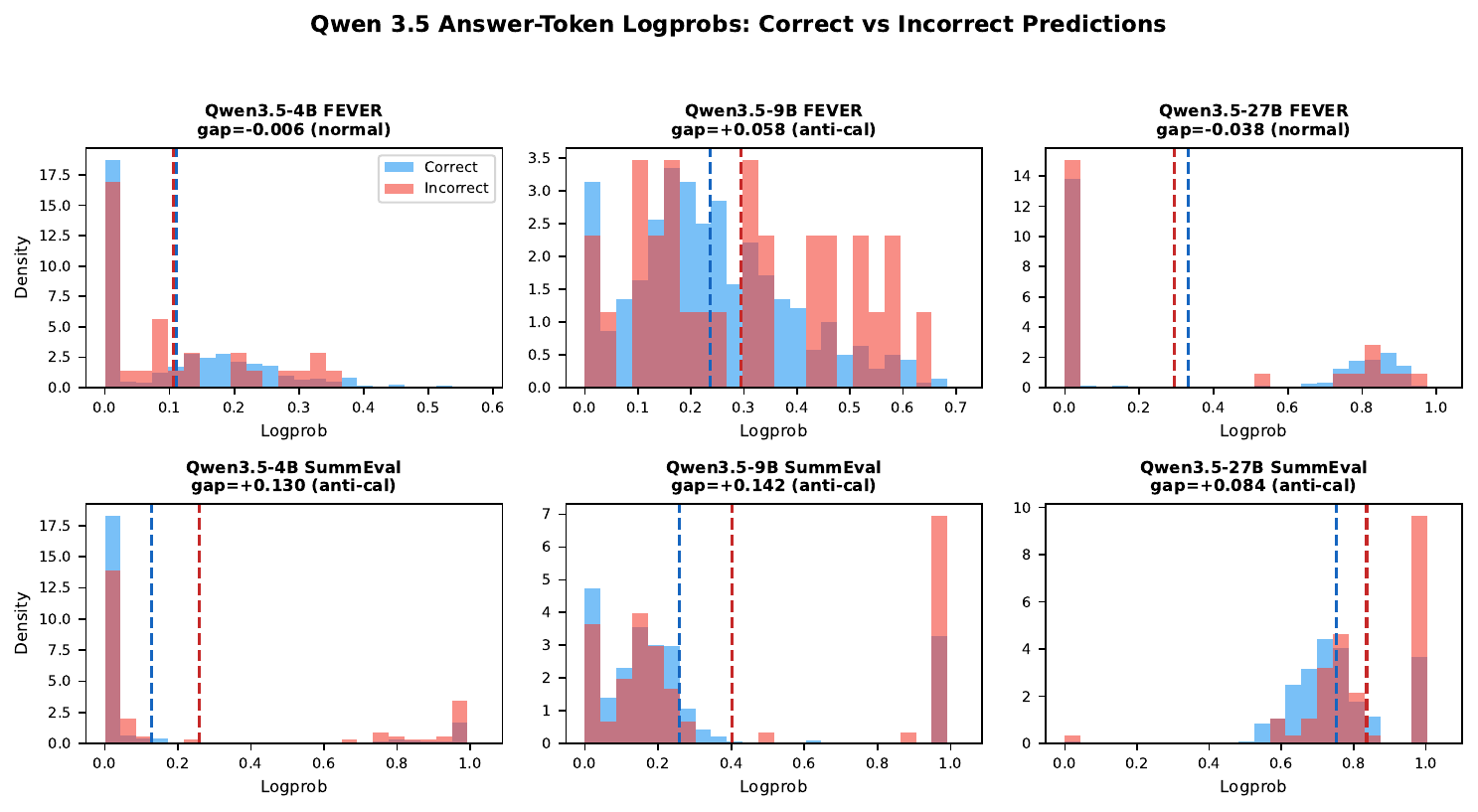}
\caption{Answer-token logprob distributions for correct (blue) vs.\ incorrect (red) predictions across three Qwen~3.5 model sizes on FEVER and SummEval. Dashed lines show means. In 4 of 6 settings, incorrect predictions have \emph{higher} mean logprobs than correct predictions (anti-calibrated), explaining the below-random \auroc{} of logprob-based confidence.}
\label{fig:antical}
\end{figure}

\section{NLI vs.\ Regex SVA Extraction}
\label{app:nli}

\begin{table}[H]
\centering
\caption{Regex vs NLI SVA extraction (\auroc{}, GPT-4.1-mini internal data).}
\label{tab:nli}
\small
\begin{tabular}{lccc}
\toprule
Rubric & SVA Regex & SVA NLI & $\Delta$ \\
\midrule
Q2 Attrib.\ Accuracy (structured) & \textbf{0.821} & 0.666 & $-$0.16 \\
Q5 Skill Relevancy (unstructured)  & 0.528 & \textbf{0.886} & +0.36 \\
Q11 Overpromising (no evidence)    & 0.504 & \textbf{0.673} & +0.17 \\
\bottomrule
\end{tabular}
\end{table}

\end{document}